\pdfoutput=1

\documentclass[11pt]{article}

\usepackage[]{acl}

\usepackage{times}
\usepackage{latexsym}
\usepackage{color, soul}
\usepackage{graphicx}
\usepackage{amsmath,amssymb}
\usepackage{diagbox}

\usepackage[T1]{fontenc}

\usepackage[utf8]{inputenc}

\usepackage{microtype}
\usepackage{booktabs}
\usepackage{multirow}

\usepackage{tikz}
\usetikzlibrary{shapes,decorations,arrows,calc,arrows.meta,fit,positioning}
\tikzset{
    -Latex,auto,node distance = 1 cm and 1 cm,semithick,
    state/.style ={ellipse, draw, minimum width = 0.7 cm},
    point/.style = {circle, draw, inner sep=0.04cm,fill,node contents={}},
    bidirected/.style={Latex-Latex,dashed},
    el/.style = {inner sep=2pt, align=left, sloped}
}
\definecolor{mygreen}{RGB}{0,153,0}
\definecolor{myred}{RGB}{153,0,0}

\title{Instructions for *ACL Proceedings}
\title{Estimating the Causal Effects of\\ Natural Logic Features in Neural NLI Models }

\author{Julia Rozanova$^{1}$,~ Marco Valentino$^{2}$,~ Andr\'{e} Freitas$^{1,2}$\\
University of Manchester, United Kingdom$^{1}$ \\
\texttt{\{firstname.lastname\}@manchester.ac.uk}$^{1}$\\
Idiap Research Institute, Switzerland$^{2}$ \\
\texttt{\{firstname.lastname\}@idiap.ch}$^{2}$\\
}
\begin{document}
\maketitle
\begin{abstract}
    Rigorous evaluation of the causal effects of semantic features on language model predictions can be hard to achieve for natural language reasoning problems. However, this is such a desirable form of analysis from both an 
    interpretability and model evaluation perspective, that it is valuable
    to zone in on specific patterns of reasoning with enough structure and regularity
    to be able to identify and quantify systematic reasoning failures in widely-used models. In this vein, we pick a portion of the NLI task for which an explicit causal diagram can be systematically constructed: in particular, the case where across two sentences (the premise and hypothesis), two related words/terms occur in a shared context.
    
    In this work, we apply causal effect estimation strategies to measure the effect of \emph{context} interventions 
    (whose effect on the entailment label is mediated by the semantic \emph{monotonicity} characteristic) and interventions on the inserted 
    word-pair (whose effect on the entailment label is mediated by the \emph{relation} between these words.).
    Following related work on causal analysis of NLP models in different settings, we 
    adapt the methodology for the NLI task to construct comparative model
    profiles in terms of \emph{robustness to irrelevant changes} and \emph{sensitivity to desired changes}.
    \end{abstract}
    
    \section{Introduction}
    
    There is an abundance of reported cases where high accuracies in NLP tasks can be attributed to simple heuristics and dataset artifacts~\cite{mccoy}.
    As such, when we expect a language model to capture a specific reasoning strategy or correctly use certain semantic features, it has become good practice to perform evaluations that provide a more granular and qualitative view into model behaviour and efficacy.
    In particular, there is a trend in recent work to incorporate causal measures
    and \emph{interventional} experimental setups in order to better understand the
    captured features and reasoning mechanisms of NLP models ~\cite{vig-gender,finlayson-agreement,stolfo-maths,geiger-causal-abstraction}.
   
    \begin{figure}[t]
        \centering
        \includegraphics[width=\columnwidth]{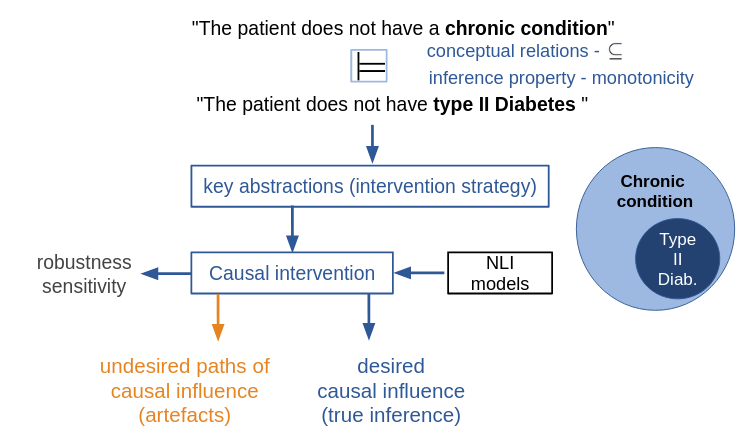}
        \caption{We propose a causal intervention framework for systematically inspecting monotonicity reasoning in NLI models.}
        \label{fig:workflow}
    \end{figure} 
    In general, it can be hard to pinpoint all the intermediate features and critical representation elements which are guiding the inference behind an NLP task. 
    However, in many cases there are subtasks which have enough semantic/logical regularity to perform stronger analyses and diagnose clear points of failure within larger tasks such as NLI and QA (Question Answering).
    As soon as we are able to draw a causal diagram which captures a portion of the model's expected reasoning capabillities, we may be guided in the design of interventional experiments which allow us to estimate 
    causal quantities of interest, giving insight into how 
    different aspects of the inputs are used by models.

    In this work, we look at a structured subset of the NLI task~\cite{rozanova-decomposing} to investigate the use of two semantic inference features by NLI models: concept
     relations and logical monotonicity.

    Our workflow is described in figure~\ref{fig:workflow}: 
    we use intermediate abstracted semantic feature labels to construct \emph{intervention sets} out of NLI examples which allow us to measure
    certain causal effects. Building upon recent work on causal analysis of NLP models~\citet{stolfo-maths}, we use 
    the ntervention sets to systematically and quantitatively characterise models' \emph{sensitivity} to relevant changes in these semantic features and \emph{robustness} to irrelevant changes.



    Our contributions may be summarised as follows:
    \begin{itemize}
        \item Following previous work on causal analysis of NLP models, we zone in on a structured subproblem in NLI (in our case, a natural logic based subtask)  and present a causal diagram which captures both desired and undesired potential reasoning routes which may describe model behaviour.
        \item We adapt the NLI-XY dataset of~\citet{rozanova-decomposing} to a meaningful collection of \emph{intervention sets} which enable the computation of certain causal effects.
        \item We calculate estimates for undesired direct causal effects and desired total causal effects, which also serve as a quantification of model robustness and sensitivity to our intermediate semantic features of interest.
        \item We compare a suite of mainstream neural NLI models, identifying behavioural weaknesses in high-performing models and behavioural advantages in some worse-performing ones.
    \end{itemize}%
    
    To the best of our knowledge, we are the first to complement previous observations of models' brittleness with respect to context monotonicity with 
    the evidence of causal effect measures, as  well as presenting new insights 
    that over-reliance on lexical relations is consequently also 
    tempered by the same improvement strategies.
    
    \section{Problem Formulation}
    
    \subsection{A Structured NLI Subtask} \label{sec:nli_subproblem}
    
    As soon as we have a concrete description of how a reasoning problem \emph{should} be treated, we can begin to evaluate how well a model emulates the expected behaviour and whether it is capturing the semantic abstractions at play. 
    
    In this work, we consider an NLI subtask which comes from the broader setting of \emph{Natural Logic} 
   ~\cite{maccartney-manning, humoss_polarity, sanchezvalencia}.
    As it has a rigid and well-understood structure, it is often used in interpretability and 
    explainability studies for NLI models~\cite{geiger-causal-abstraction, richardson_fragments, geiger-inducing, rozanova-decomposing, rozanova_supporting}. We begin with the format described in~\cite{rozanova-decomposing} 
    (we refer to this work for more detailed description and full definitions). 
    
    \noindent Consider two terms/concepts with a known \emph{relation label}, 
    such as one of the pairs: 
    
    \begin{table}[h!]
        \centering
        \begin{tabular}{ccc}
        \midrule
        Word/Term $x$ & Word/Term $y$  & Relation  \\
        \midrule
        brown sugar & sugar & $x \sqsubseteq y$ \\
        mammal & lion & $x \sqsupseteq y$ \\
        computer & pomegranate & $x \# y$ \\
        \midrule
    \end{tabular}
        \caption{Examples of the word-pair/term-pair relation informing entailment examples.
        Concept inclusion and its dual are related by $\sqsubseteq$ and $\sqsupseteq$ respectively,
        while $\#$ denotes the absence of either relations.}
    \end{table}
    
    \noindent  Suppose the two terms occur in an identical context (comprising of a natural language sentence, like a template), for example:
    
    \begin{table}[h!]
        \centering
        \begin{tabular}{ll}
        \midrule
             Premise & I do not have any \textbf{sugar}.  \\
             Hypothesis & I do not have any \textbf{brown sugar}. \\\midrule
        \end{tabular}
    \end{table}
    
    A semantic property of the natural language context called \emph{monotonicity} determines whether there is an \emph{entailment 
    relation} between the sentences generated upon substitution/insertion
    of given related terms (formally, this is monotonicity in the sense 
    of preserving the ``order" between the inserted terms to an equally-directed entailment relation between the sentences.) 
    The context monotonicity may either be \emph{upward} ($\uparrow$) or
    \emph{downward} ($\downarrow$, as in the example above) or \emph{neither}. 
    
\begin{figure}
        \centering
        \begin{tabular}{cc}
             Variable & Description \\ \midrule
             $G$ & Gold Label\\
             $C$ &  Context \\
             $M$ &  Context Monotonicity\\
             $W$ &  Inserted Word Pair \\
             $R$ &  Word-Pair Relation \\
        \end{tabular}
       
        \vspace{1em}
        
        \begin{tikzpicture}[scale=1.5]
            \node[state] (c) at (0,2) {$C$};
            \node[state] (m) at (1,2) {$M$};
            
            \node[state] (w) at (0,1) {$W$};
            \node[state] (r) at (1,1) {$R$};
            \node[state] (g) at (2,1.5) {$G$};
    
            \path (c) edge (m);
            \path (m) edge[bend left=20] (g);
    
            \path (w) edge (r);
            \path (r) edge[bend right=20] (g);
            
        \end{tikzpicture}
        \caption{Causal Diagram for the Natural Logic Subtask}
        \label{fig:nli_xy_causal_diagram}
    \end{figure}
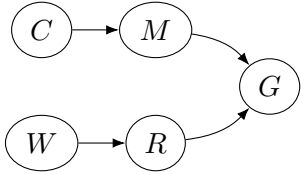

    We can describe how the monotonicity behaves in conjunction with the relation between the inserted words in the causal diagram in figure~\ref{fig:nli_xy_causal_diagram}.
    
    The way in which the value of these variables affects the gold entailment label is summarised in table~\ref{table:value_summary}.
    
    \begin{table}[h!]
        \centering
        \resizebox{0.7\columnwidth}{!}{
        \begin{tabular}{c|ccc}
        \backslashbox{$M$}{$R$} & $\sqsubseteq$ & $\sqsupseteq$ & $\#$ \\ \hline
             $\uparrow$ & Entailment & Non-Entailment &Non-Entailment \\ 
             $\downarrow$ &Non-Entailment& Entailment& Non-Entailment \\ 
        \end{tabular}%
        }
        \caption{The entailment gold labels as a function of two semantic features: the context montonicity (M) and the relation (R) of the inserted word pair.}
        \label{table:value_summary}
    \end{table}
    The authors of~\citet{rozanova-decomposing} provide a thus-formatted dataset called \emph{NLI-XY}, which we use as the basis for our causal effect estimation experiments.
    
    Naturally, it is always likely that models may fail to follow the described reasoning scheme for these NLI problems. In the next section (\ref{sec:model_causal_diagram}), we propose a causal diagram
    which also captures the reasoning possibilities an NLI model may follow, accounting for possible 
    confounding heuristics via unwanted direct effects. 
    
    \subsection{The Causal Structure of Model Decision-Making}\label{sec:model_causal_diagram}
    
    In an ideal situation, a strong NLI model would identify the word-pair relation and the context 
    monotonicity as the abstract variables relevant to the final entailment label.
    In this case, these features would causally affect the model prediction in the same way 
    they affect the gold label. 
    Realistically, as shown in illuminating studies such as~\citet{mccoy-right-for-the-wrong},
    models identify unexpected biases in the dataset and may end up using accidental 
    correlations output labels, such as the frequency of certain words in a corpus.
    For example,~\citet{mccoy-non-entailment} demonstrate how models can successfully exploit 
    the presence of negation markers to anticipate non-entailment, even when it is not 
    semantically relevant to the output label.
    
    To ensure that the semantic features themselves are taken account into the model's output and not 
    other surface-level confounding variables, one would like to perform interventional studies
    which alter the value of the target feature but not other confounding variables.
    This is, in many cases, not feasible (although attempts are sometimes made to at least perform 
    interventions that only make minimal changes to the surface form, as in~\citet{kaushik-counterfactually-augmented}.)
    
   ~\citet{stolfo-maths} argue that it is useful to quantify instead the direct impact of irrelevant 
    surface changes (controlling for values of semantic variables of interest) and compare them to 
    \emph{total causal effects} of input-level changes: doing so, we may posit deductions about the 
    flow of information via the semantic variables (or lack thereof).
    For analyses where there is an attempt to align intermediate variables with explicit internals,
    see vig-gender,~\citet{finlayson-agreement} for a mediation analysis approach, or~\citet{geiger-causal-abstraction} for an alignment strategy based on causal abstraction theory.
    
    \paragraph{Diagram Specification}
    We follow~\citet{stolfo-maths} in the strategy of explicitly modeling the
    ``irrelevant surface form" of the input text portions as variables in the causal diagram.
    Their setting of \emph{math word problems} is decomposed into two compositional inputs: 
    a question template and two integer arguments.
    Our setting follows much the same structure: our natural language ``context'' plays the same role as 
    their ``template'', but our arguments (an inserted word pair) have an additional layer of complexity as we also model the 
    \emph{relation} between the arguments as an intermediate reasoning variable rather than 
    the values themselves (as such, the structure of their template modeling in their causal diagram
    is more applicable than the direct way they treat their numerical arguments.)
    
    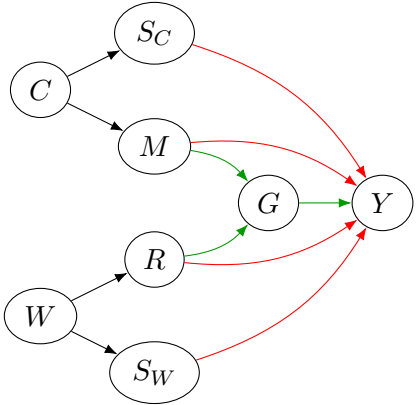
\begin{figure}[t]
        \centering
        \begin{tabular}{cc}
             Variable & Description \\ \midrule
             $Y$ & Model Prediction\\
             $G$ & Gold Label\\
             $C$ &  Context \\
             $M$ &  Context Monotonicity\\
             $S_{C}$ &  Context Textual Surface Form\\
             $W$ &  Inserted Word Pair \\
             $R$ &  Word-Pair Relation \\
             $S_{W}$ &  Word-Pair Textual Surface Form\\
        \end{tabular}
        \vspace{1em}
        
        \begin{tikzpicture}[scale=1.5]
            \node[state] (c) at (0,2.5) {$C$};
            \node[state] (sc) at (1,3) {$S_{C}$};
            \node[state] (m) at (1,2) {$M$};
            
            \node[state] (w) at (0,0.5) {$W$};
            \node[state] (sw) at (1,0) {$S_{W}$};
            \node[state] (r) at (1,1) {$R$};
            \node[state] (g) at (2,1.5) {$G$};
            \node[state] (y) at (3,1.5) {$Y$};
    
            \path (c) edge (sc);
            \path (c) edge (m);
            \path[red] (sc) edge[bend left=20] (y);
            \path[mygreen] (m) edge[bend left=20] (g);
            \path[red] (m) edge[bend left=20] (y);
    
            \path (w) edge (sw);
            \path (w) edge (r);
            \path[red] (sw) edge[bend right=20] (y);
            \path[mygreen] (r) edge[bend right=20] (g);
            \path[red] (r) edge[bend right=20] (y);
            \path[mygreen] (g) edge (y);
            
        \end{tikzpicture}
        \caption{Specification of the causal diagram for possible routes of model reasoning for NLI-XY problems. 
        Green edges indicate \emph{desired} causal influence, while red edges indicate \emph{undesired} paths of causal influence via surface-level heuristics.} 
        \label{fig:causal_diagram}
    \end{figure}
    
    We present our own causal diagram in figure~\ref{fig:causal_diagram}. We introduce the textual context
    $C$ as an input variable, which is further decomposed into more abstract variables: 
    its \emph{monotonicity} $M$ (which directly affects the gold truth $G$) and the textual surface form
    of the context $S_{C}$.
    The other input variable is the word-pair insertion which we will summarise as a single variable $W$. Once again, $W$ has a potential effect on the model decision through its surface form $S_{W}$ and the via relation $R$ between the words. The gold truth $G$ is dependant on $W$ and $R$ only.  
    Finally, the outcome variable is the model prediction $Y$.
    The paths for which we would like to observe the highest causal effect are the paths to $Y$ from the inputs via $M, R$ and through the gold truth variable $G$.
    However, each of $S_{C}, S_{W}, M, R, C$ and $W$ have direct links to the model output $Y$ as well (indicated in red):  these are potential direct effects which are \emph{unwanted}.  
    The key goal of this study is to compare the extent to which models exhibit the high causal effects for the \emph{desired} diagram routes and lower causal effects for the \emph{undesired} routes.
    
    \section{Estimating the Causal Effects}
    Causal effects can be calculated via the implementation of interventional experiments.
    The calculation strategy for the causal effect of a given \emph{source} variable on a 
    \emph{target} variable depends on the structure of the causal diagram, and
    subsequently the set of interventions and controls that need to be performed in
    order to estimate the causal influence in question.
    Only certain interventions are feasible in our setting, so we mention right off the 
    bat that most of the causal effects that are of greatest interest will not be possible to estimate. 
    However, as with the work in~\citet{stolfo-maths}, there are enough useful interventions we can perform (and thus causal effects we can measure) to 
    make useful model comparisons and develop some insights into models' robustness and
    sensitivity to distinct classes of of input changes.
    
    For the algebra of causal effect calculation, we draw heavily on the reasoning in~\citet{stolfo-maths}, as the upper portion of their causal diagram 
    (concerning the natural language question template, its surface form and the implicit math operation) is equivalent to both the upper and lower half of our diagram in figure~\ref{fig:causal_diagram}.
    
    We start with the \emph{Total Causal Effect} (TCE) of the respective textual inputs $C$ (the context) and $W$ (the inserted word pair).
    
    \paragraph{Total Causal Effects} 
    To obtain the total causal effect of the context $C$, we intervene on the context \emph{without conditioning on the context monotonicity.}
    It can  calculated as
    \begin{align}
    \mbox{TCE}(C ~\mbox{on } Y) &:= \mathbb{E}^{\textrm{int}+}_{C}[Y] - 
    \mathbb{E}^{\textrm{int}-}_{C}[Y],
    \end{align}
    
    where $\mathbb{E}^{\textrm{int}+}_{C}[Y]$ indicates the expected 
    prediction value \emph{after} intervention on $C$, while $\mathbb{E}^{\textrm{int}-}_{C}[Y]$ denotes the expected outcome \emph{before}
    this intervention.
    
    On the other hand, to obtain the total causal effect of the inserted word pair $W$, we intervene on the word pair \emph{without conditioning on the context monotonicity.}
    It can  calculated as
    \begin{align}
    \mbox{TCE}(W ~\mbox{on } Y) &:= \mathbb{E}^{\textrm{int}+}_{W}[Y] - 
    \mathbb{E}^{\textrm{int}-}_{W}[Y].
    \end{align}
    
    This is a quantification of their respective causal influence on the outcome $Y$
    \emph{through all possible paths}: in particular, this captures both the undesired route of influence (via surface heuristics)
    and the desired route (via the corresponding semantic feature and gold truth label).
    
    \begin{table*}[htb!]
    \resizebox{\textwidth}{!}{
    \begin{tabular}{@{}llp{5cm}p{5cm}lll@{}}
    \toprule
    Target Quantity                     & Intervention & Premise                       & Hypothesis                            & M            & R             & G              \\ \midrule
    \multirow{2}{*}{TCE($W \to Y$)}     & Before       & There's a cat on the pc.   & There's a cat on the machine.      & $\uparrow$   & $\sqsubseteq$ & Entailment     \\
                                        & After        & There's a cat on the tree. & There's a cat on the fruit tree.     & $\uparrow$   & $\sqsupseteq$ & Non-Entailment \\ \midrule
    \multirow{2}{*}{DCE($S_{W} \to Y$)} & Before       & There are no students yet.    & There are no first-year students yet. & $\downarrow$ & $\sqsupseteq$ & Entailment     \\
                                        & After        & There are no people yet.     & There are no women yet.                & $\downarrow$ & $\sqsupseteq$ & Entailment     \\ \midrule
    \end{tabular}%
    }
    \caption{Example word-pair insertion interventions for determining the total causal effect of label-relevant word-pair changes and the direct causal effect of
    label-irrelevant word-pair changes.}
    \label{table:insertion_int_examples}
    \end{table*}
    
    \begin{table*}[htb!]
    \resizebox{\textwidth}{!}{
    \begin{tabular}{@{}llp{5cm}p{5cm}lll@{}}
    \toprule
    Target Quantity                     & Intervention & Premise                                                & Hypothesis                                             & M            & R            & G              \\ \midrule
    \multirow{2}{*}{TCE($C \to Y$)}     & Before       & You can't live without fruit .                       & You can't live without strawberries .                & $\uparrow$   & $\sqsupseteq$ & Non-Entailment \\
    
                                        & After        & All fruit study english.                              & All strawberries study English.                       & $\downarrow$ & $\sqsupseteq$ & Entailment     \\ \midrule
    \multirow{2}{*}{DCE($S_{C} \to Y$)} & Before       & He has no interest in seafood .                        & He has no interest in oysters .                        & $\downarrow$ & $\sqsupseteq$ & Entailment     \\
                                        & After        & I don't want to argue about this in front of seafood . & I don't want to argue about this in front of oysters . & $\downarrow$ & $\sqsupseteq$ & Entailment     \\ \bottomrule
    \end{tabular}%
    }
    \caption{Example context interventions for determining the total causal effect of label-relevant context changes and the direct causal effect of
    label-irrelevant context changes.}
    \label{table:context_int_examples}
    \end{table*}

    \paragraph{Undesired Direct Causal Effects}
    The key suggestion in~\citet{stolfo-maths} is that even though we have no feasible intervention 
    strategies which allow us to calculate the direct causal effect (DCE) of the intermediate variable 
    $M$ (respectively, $R$), we may yield some insight into their causal influence by comparing the 
    relevant TCE to the \emph{unwanted direct effect} DCE ($S_C \to Y$) (respectively, DCE ($S_W \to Y$)).
    
    In principle, this direct effect can be seen as \emph{accounting for a portion of the total causal 
    influence} which is indicated by the TCE.  More importantly, this quantity can serve as an estimate of \emph{robustness} to irrelevant changes in
    the given input variable.
    We calculate it by conditioning on the corresponding semantic feature variables of interest (note that for these interventions, the gold label is unchanged as it is
    dependant on the variables we are conditioning for):
    
    \begin{equation}
    \begin{split}
    &\mbox{DCE}(S_C ~\mbox{on } Y) := \\
    &\sum_m P(M) (\mathbb{E}^{\textrm{int}+}_{S_C}[Y\mid M=m] - \mathbb{E}^{\textrm{int}-}_{S_C}[Y\mid M=m])\\ 
    \end{split}
    \end{equation}
    \begin{equation}
    \begin{split}
    &\mbox{DCE}(S_W ~\mbox{on } Y) := \\
    &\sum_r P(R) (\mathbb{E}^{\textrm{int}+}_{S_W}[Y\mid R=r] - \mathbb{E}^{\textrm{int}-}_{S_W}[Y\mid R=r]).\\ 
    \end{split}
    \end{equation}
    
    \section{Experimental Setup}
    \subsection{Data and Interventions}\label{sec:intervention_scheme}
    Whether we are measuring a DCE or TCE differs mostly in the choice of interventions. As such, the core of this work is compiling the relevant data into meaningful sets of interventions.
    To this end, let $N$ denote an NLI-XY example consisting of a tuple $(P, H, C, W, M, R, G)$. If we allow that $W$ is a pair $(w_1, w_2)$, the premise $P$ and hypothesis $H$ are substitutions into the context $C$, so that $P=C(w_1)$, $H=C(w_2)$. The gold label $G$ is determined as by
    table~\ref{table:value_summary} depending on the values of $M$ and $R$.

    We define an \emph{intervention} as a pair $(N, N')$ of NLI-XY examples, which will be subject
    to certain conditioning restrictions depending on the causal effect quantity we wish to determine.
    These conditioning restrictions are summarized by the \emph{intervention schema} in table~\ref{table:intervention_schema}.
    
    \begin{table}[htb!]
        \centering
    \resizebox{0.6\columnwidth}{!}{
    \begin{tabular}{@{}lllllllp{2.5cm}@{}}
    \toprule
    Target Measure& $C$ & $W$ & $M$ & $R$ & $G$ & & Interventions in Dataset\\ \midrule
    DCE $(S_C \to Y)$ & $\not =$ & $=$ & $=$ &$=$ & $=$ & & 20910\\
    TCE $(C \to Y)$ &$\not=$ & $=$ & $\not =$ & $=$ & $\not=$ & & 14270\\
    \bottomrule
    DCE $(S_W \to Y)$ & $=$ & $\not =$ & $=$ & $=$ & $=$ & & 25960\\
    TCE $(W \to Y)$ & $=$ & $\not=$ & $=$ & $\not =$ & $\not =$ & &  22640\\
    \midrule
    \end{tabular}%
    }
    \caption{Intervention schema and dataset statistics: which variables are held constant and which are changed
    in the construction of intervention sets for the calculation of the indicated effects.}
    \label{table:intervention_schema}
    \end{table}

    We use the NLI-XY evaluation dataset to construct intervention pairs $(N,N')$ by using a sampling/filtering strategy 
    as in~\cite{stolfo-maths} according to the 
    intervention schema in table~\ref{table:intervention_schema}.
    In particular, for constructing \emph{context} interventions, we sample a seed set of 400 NLI-XY
    premise/hypothesis pairs. This is the \emph{pre-intervention} NLI example. 
    For each, we fix the insertion pair and filter through the NL-XY dataset for all 
    examples with the shared insertion pair but different context, conditioned as necessary 
    on the properties of the other variables as in the intervention schema. For insertion pairs, we do the opposite.
    The number of interventions we produce in this way for our experiments are reflected in the last column of 
    table~\ref{table:intervention_schema}

    In summary, the changes are context replacements and related word-pair replacements; 
    we provide text-level examples in tables~\ref{table:insertion_int_examples} and~\ref{table:context_int_examples}.

    \subsection{Model Choice and Benchmark Comparison}\label{sec:models}
    
    We include the following models \footnote{All pretrained models are from the Huggingface \emph{transformers} library (\cite{transformers}),
    except for \textbf{infobert} and the pretrained model counterparts fine-tuned on \textbf{HELP}: their sources are linked in the README of the accompanying code.} in our study:
    
    \begin{itemize}
        \item The models evaluated in NLI-XY paper~\cite{rozanova-decomposing}, namely
        \textbf{roberta-large-mnli}, \textbf{facebook/bart-large-mnli}, \textbf{bert-base-uncased-snli} and their counterparts fine-tuned on the \textbf{HELP} dataset~\cite{yanaka-help}
        \item The \textbf{infobert} model, which is trained on three benchmark training
        sets of interest: \textbf{MNLI}~\cite{mnli}, \textbf{SNLI}~\cite{snli} and \textbf{ANLI}~\cite{anli} (currently at the top of the leaderboard for the adversarial ANLI test set, as of January 2023)
        \item Another \textbf{roberta-large} checkpoint, also trained on all three benchmark NLI training training sets (as well as FEVER-NLI~\cite{fever}).
    \end{itemize}
    We report their scores on the mentioned benchmark datasets alongside 
    the relevant total and direct causal effects we are interested in.
    
    Note that as the HELP dataset is a two class entailment dataset (as opposed to datasets like MNLI, which 
    are three class), we cannot directly compare existing reported scores. 
    As such, we adapt the three class scores to a two class score by grouping two of the three class labels 
    (``contradiction" and ``neutral") into the two class umbrella label "non-entailment".
    For all models, we report both the three class and adapted two class accuracy scores on the benchmark 
    datasets.

    
    \section{Results and Discussion}
    We examine and compare the results for the models listed in~\ref{sec:models}.
    We first zoom in on the word-pair insertion intervention experiments in~\ref{sec:insertion_results}, then the context intervention experiments in~\ref{sec:context_results} and finally present a categorical overview of these results in section~\ref{sec:comparison_results}, contextualised by benchmark scores.
    
    \subsection{Causal Effect of Inserted Word Pairs} \label{sec:insertion_results}
\begin{figure}[h!]
        \centering
        \includegraphics[width=\columnwidth]{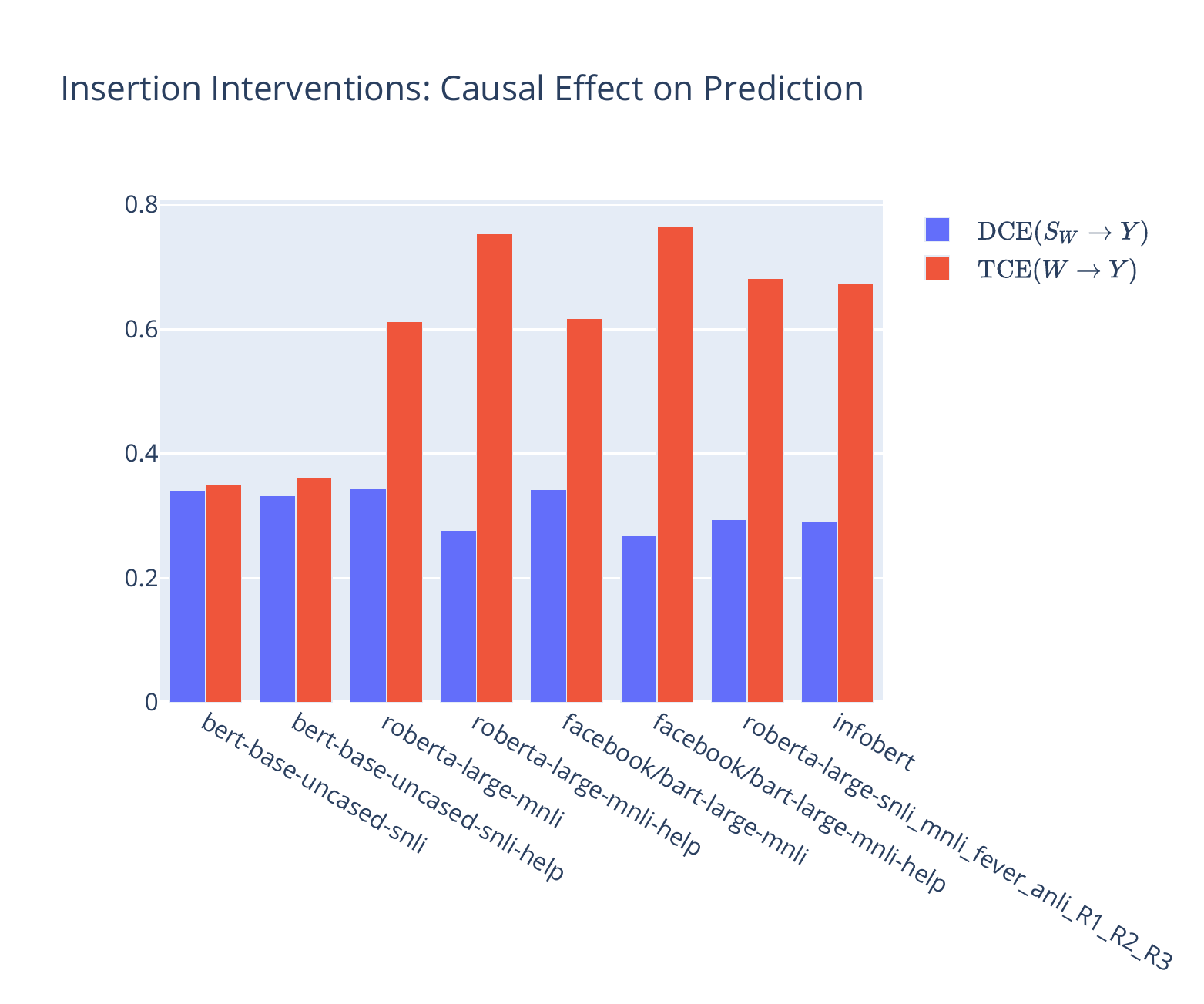}
    \resizebox{\columnwidth}{!}{
    \begin{tabular}{@{}p{5.5cm}lrrlrr@{}}
    \toprule
    Model                                             &  & \multicolumn{1}{l}{DCE($S_W \to Y$)} & \multicolumn{1}{l}{TCE($W \to Y$)} &  & \multicolumn{1}{l}{TCE/DCE Ratio} & \multicolumn{1}{l}{Delta} \\ \midrule
    bert-base-uncased-snli                            &  & 0.341                                             & 0.350                                          &  & 1.027                          & 0.009                  \\
    bert-base-uncased-snli-help                       &  & 0.332                                             & 0.361                                          &  & 1.087                           & 0.029                  \\
    roberta-large-mnli                                &  & 0.343                                             & 0.613                                          &  & 1.785                          & 0.269                  \\
    roberta-large-mnli-help                           &  & 0.276                                             & 0.754                                          &  & 2.730                          & 0.478                  \\
    facebook/bart-large-mnli                          &  & 0.342                                             & 0.618                                          &  & 1.805                          & 0.275                  \\
    facebook/bart-large-mnli-help                     &  & 0.268                                             & 0.766                                          &  & 2.863                          & 0.499                  \\
    roberta-large-snli\_mnli\_fever\_anli\_R1\_R2\_R3 &  & 0.294                                             & 0.682                                            &  & 2.321                          & 0.388                  \\
    infobert                                          &  & 0.291                                             & 0.674                                          &  & 2.320                          & 0.384                  \\ \bottomrule
    \end{tabular}%
    }
        \caption{Results for Insertion Interventions}
        \label{fig:insertion_experiment}
    \end{figure}

    The results for the substituted word-pair intervention experiment are reported in
    figure~\ref{fig:insertion_experiment}.
    The most desireable outcome is a DCE($S_C \to Y$) which is \emph{as low as possible}
    in combination with a TCE($C \to Y$) which is \emph{as high as possible}.
    The lower this DCE, the higher the model robustness to \emph{irrelevant context surface form} changes. On the other hand, the higher the specified TCE, the greater
    the model's sensitivity to \emph{context changes affecting the gold label}. 
    
    The largest delta between these two quantities can be seen in the 
    \textbf{roberta-large-mnli-help} and \textbf{facebook-bart-large-mnli-help} models.
    This is important to note: the HELP dataset~\cite{yanaka-help} is explicitly designed 
    to bolster model success on natural logic problems, but until now there has been 
    little to no evidence that it improves the treatment of word-pair relations.
    In particular, the internal probing results in~\cite{rozanova-decomposing} show that
    probing performance for the intermediate word-pair relation label decreases slightly
    for \textbf{roberta-large-mnli} after fine-tuning on HELP; as such, it was thought
    that the HELP improvements on natural logic could solely be attributed to improved 
    context monotonicity treatment.
    Now, however, we observe distinct improvements in robustness to irrelevant word-pair 
    insertion changes and sensitivity to relevant ones.
    
    More generally, the work in~\citet{rozanova-decomposing} does indicate that the large MNLI-based 
    models are already very successful in distinguishing the relation between substituted words.
    The word-pair relation label has a high \emph{probing} result for all of these models, as well as strong signs of systematicity in their error analysis. 
    This is in line with our observations of relatively large deltas between the DCE and TCE here, compared to the smaller BERT-based models.  
    
    \subsection{Causal Effect of Contexts}\label{sec:context_results}
\begin{figure}[h!]
        \centering
        \includegraphics[width=\columnwidth]{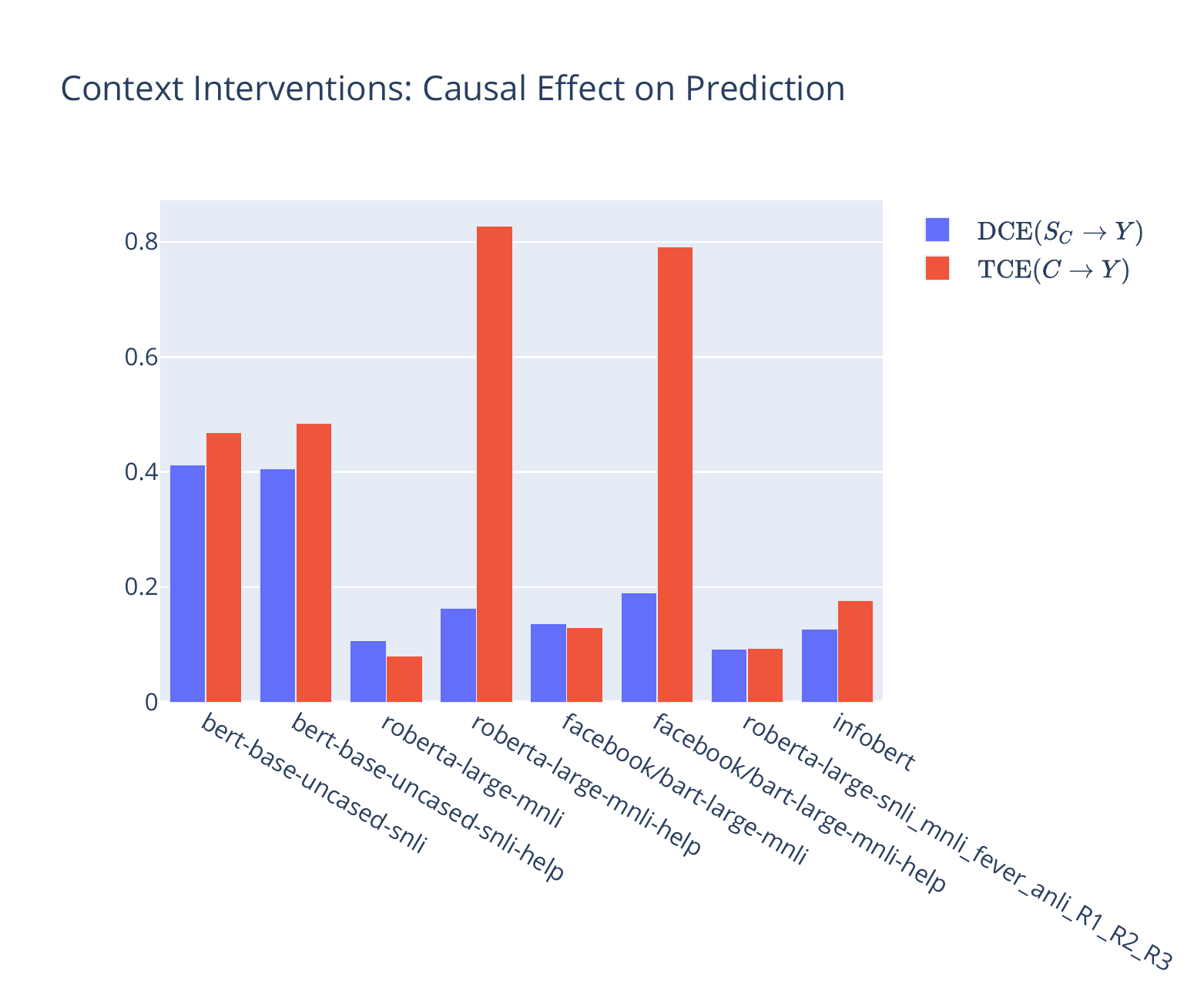}
        \resizebox{\columnwidth}{!}{
    \begin{tabular}{@{}p{5.5cm}lrrlrr@{}}
    \toprule
    Model                                             &  & \multicolumn{1}{l}{DCE($S_C \to Y$)} & \multicolumn{1}{l}{TCE($C \to Y$)} &  & \multicolumn{1}{l}{TCE/DCE Ratio} & \multicolumn{1}{l}{Delta} \\ \midrule
    bert-base-uncased-snli                            &  & 0.412                             & 0.468                           &  & 1.136                           & 0.056                  \\
    bert-base-uncased-snli-help                       &  & 0.406                             & 0.485                            &  & 1.194                           & 0.079                  \\
    roberta-large-mnli                                &  & 0.107                             & 0.081                           &  & 0.751                          & -0.027                 \\
    roberta-large-mnli-help                           &  & 0.163                             & 0.828                           &  & 5.070                          & 0.665                  \\
    facebook/bart-large-mnli                          &  & 0.136                              & 0.130                           &  & 0.954                          & -0.006                 \\
    facebook/bart-large-mnli-help                     &  & 0.189                             & 0.791                           &  & 4.167                          & 0.601                  \\
    roberta-large-snli\_mnli\_fever\_anli\_R1\_R2\_R3 &  & 0.093                             & 0.093                           &  & 1.008                          & 0.001                   \\
    infobert                                          &  & 0.127                             & 0.176                           &  & 1.385                           & 0.049                  \\ \bottomrule
    \end{tabular}%
    }
        \caption{Results for Context Interventions}
        \label{fig:context_experiment}
    \end{figure}

    The results for the context intervention experiments are reported in
    figure~\ref{fig:context_experiment}.
    The most desireable outcome is a DCE($S_C \to Y$) which is \emph{as low as possible}
    in combination with a TCE($C \to Y$) which is \emph{as high as possible}.
    
    For context interventions, we start to see major distinctions in the sensitivity of models to 
    important context changes - especially the effect of the \textbf{HELP} fine-tuning 
    dataset in increasing model reasoning with respect to context structure.
    In line with previous behavioural findings in~\citet{richardson_fragments, yanaka-help, yanaka-med, geiger, rozanova-decomposing} and all the way back to~\citet{glue}, which observe systematic
    failure of large language models in downward monotone contexts, we notice that all
    of the models trained only on the large benchmarks sets fail to correctly change their prediction when a context change requires it to do so 
    (as indicated by the low TCE score). 
    
    In~\citet{yanaka-help},~\citet{rozanova-decomposing} and~\citet{rozanova_supporting},  the positive effect of the \textbf{HELP} dataset is already evident, 
    but here we may also compare it to roberta-large-mnli tuned on many additional training sets, precluding the possibility that its 
    helpfulness can be attributed only to a ``larger amount of training data".   
    
    We note that although the situation of the TCE/DCE ratio for 
    \textbf{roberta-large-mnli} being less than one may seem peculiar, it is important to 
    keep in mind that the intervention sets used for estimating these quantities are sampled independently so some margin of error is warranted.
    As in~\citet{stolfo-maths}, we interpret this result to simply mean that the 
    causal influence is comparable whether we are affecting the ground truth result (as in the TCE($C \to Y$) case) or not (as in the DCE($S_C \to Y$) case).

    \subsection{Other Potential Direct Effects}
    Note that there are some unwanted direct links whose causal influence we have not estimated in this work: in particular, the direct effects of the
    context monotonicity $M$ and the word-pair relation $R$ that do \textit{not} factor
    through the entailment gold label $G$.
    Such an instance is easy to imagine - for example, a model may be biased to 
    conclude that any instance of a correctly identified $\sqsubseteq$ related word pair across the example (hyponym in the premise, hypernym in the hypothesis) 
    should be labelled for entailment (regardless of the context, shared or not).
    In fact, this is indeed what we hypothesize the more context-monotonicity-ignorant models (such as roberta-large-mnli) are doing.
    
    \subsection{Benchmark Scores and Causal Effects}\label{sec:comparison_results}

    \begin{table*}[]
    \resizebox{\textwidth}{!}{
    \begin{tabular}{@{}lllllllllllllll@{}}
    \toprule
    \multicolumn{1}{c}{\multirow{2}{*}{\textbf{Model}}} &  & \multicolumn{6}{l}{NLI Benchmark Evaluation (2 Class Accuracy)}                                     &  & \multicolumn{2}{l}{Context Changes} &  & \multicolumn{2}{l}{Inserted Word-Pair Changes} &  \\ \cmidrule(lr){2-8} \cmidrule(lr){10-11} \cmidrule(lr){13-14}
    \multicolumn{1}{c}{}                                &  & SNLI           & MNLI-M         & MNLI-MM        & ANLI-R1        & ANLI-R2        & ANLI-R3        &  & Robustness       & Sensitivity      &  & Robustness             & Sensitivity           &  \\ \midrule
    bert-base-uncased-snli                              &  & 0.766          & 0.620          & 0.623          & 0.567          & 0.596          & 0.580          &  & Mid              & Mid              &  & Mid                    & Low                   &  \\
    bert-base-uncased-snli-help                         &  & 0.757          & 0.627          & 0.626          & 0.505          & 0.508          & 0.546          &  & Mid              & Mid              &  & Mid                    & Low                   &  \\
    facebook/bart-large-mnli                            &  & 0.935          & 0.940          & 0.939          & 0.596          & 0.563          & 0.593          &  & High             & Low              &  & Mid                    & Mid                   &  \\
    facebook/bart-large-mnli-help                       &  & 0.727          & 0.802          & 0.795          & 0.538          & 0.489          & 0.528          &  & Mid/High         & \textbf{Highest} &  & \textbf{Highest}       & \textbf{Highest}      &  \\
    roberta-large-mnli                                  &  & 0.931          & 0.941          & 0.940          & 0.614          & 0.529          & 0.5325         &  & \textbf{Highest} & Lowest           &  & Mid                    & Mid                   &  \\
    roberta-large-mnli-help                             &  & 0.738          & 0.668          & 0.656          & 0.565          & 0.554          & 0.574          &  & High             & \textbf{Highest} &  & \textbf{Highest}       & \textbf{Highest}      &  \\
    roberta-large-snli\_mnli\_fever\_anli   &  & 0.949          & 0.936          & 0.939          & 0.810          & 0.659          & 0.666          &  & \textbf{Highest} & Lowest           &  & Mid                    & Mid/High              &  \\
    infobert                                            &  & \textbf{0.950} & \textbf{0.943} & \textbf{0.941} & \textbf{0.837} & \textbf{0.682} & \textbf{0.683} &  & High             & Low              &  & Mid                    & Mid/High              &  \\ \bottomrule
    \end{tabular}%
    }
    \caption{Overall 2 class accuracy on original NLI benchmarks and qualitative comparison against the performed causal intervention analysis. The accuracy is not necessarily predictive of the performances achieved using a systematic causal inspection.}
    \label{table:comparison_table}
    \end{table*}

    A summary of the performance of all models on popular benchmarks
    alongside a categorical breakdown of robustness and sensitivity is presented in table~\ref{table:comparison_table}. 
    The robustness/sensitivity categories are a qualitative assessment, identifying the \emph{lowest} and \emph{highest} scores within a category, and categorising other models correspondingly as \emph{low}, \emph{mid} or \emph{high} performers for the given categories. The sensitivity property is tied to the desired total causal effect, while the robustness property is tied to the undesired direct causal effect (note in particular that the latter is judged as \emph{inversely proportional:} the model with the lowest given DCE is judged the ``highest" in terms of robustness).
    
    The key observation is that the models which achieve the highest 
    performance on benchmarks may be far from the best performers with 
    respect to our quantitative markers of strong reliance of important
    causal features.
    In particular, models such as \emph{infobert} are outperformed in 
    our behavioural causal effect analyses by weaker models that are
    fine-tuned on a relatively small helper dataset such as \emph{HELP}. 
    It is important to note that such changes coincide with drops in benchmarks performance too, 
    but any model interventions that discourage the exploitation of heuristics (evident from a lower DCE for surface form features) may have that effect. 
    
    
    \section{Related Work}
    \paragraph{Natural Logic Handling in NLI Models}
    It has been known for some time that large NLI models are frequently 
    tripped up by downward-monotone reasoning~\cite{richardson-fragments, glue, yanaka-help, rozanova-decomposing, geiger-partially}.
    Various datasets have been created to evaluate and improve this behaviour, such as 
    HELP~\cite{yanaka-help}, MoNLI~\cite{geiger-partially}, MQNLI~\cite{geiger_posing_fair}, MED~\cite{yanaka-med}. 
   ~\citet{rozanova-decomposing} introduced NLI-XY, secondary compositional dataset built from portions of MED, where the intermediate features of \emph{context monotonicity} and \emph{concept relations} are explicitly 
    labelled: this is the dataset we use in this work. 
    
    Non-causal structural analyses of model internals with respect to natural 
    logic features include~\cite{rozanova-decomposing} (a probing study), but we leave to the next section some existing works where natural logic intersects 
    with the world of causal approaches to NLP. 
    
    \paragraph{Causal Analysis in NLP}
    Causal modelling has appeared in NLP works in various forms, such as the 
    investigations of the causal influence of data statistics~\cite{elazar_causal_data} and 
    mediation analyses~\cite{vig-gender, finlayson-agreement} which link intermediate linguistic/semantic features to model internals.
   ~\citet{stolfo-maths}, our core reference, appears to be the first to 
    use explicitly causal effect measures as indicators of 
    sensitivity and robustness (for some non-causal approaches to measuring model robustness in NLP, we point to~\cite{textfooler} and~\cite{checklist}).
    For a fuller summary of the use of causality in NLP, please see the 
    survey by~\citet{feder-causal-nlp}.
    
    Specific to natural logic, works with causal approaches include~\citet{geiger-partially} (which perform interchange interventions at a token representation level),~\citet{geiger-causal-abstraction} (where an ambitious causal abstraction experiment attempts to align model internals with candidate causal models) and the works of~\citet{geiger-partially} and~\cite{wu-proxy}, (where attempts are made to build a prescribed causal structure into models themselves). In particular,~\cite{wu-proxy} create a  ``causal proxy model" which becomes the basis for a new explainable predictor designed to replace the original neural network. 
    \section{Conclusion}
    The results here strongly bolster the fact that similar benchmark 
    accuracy scores may be observed for models that exhibit very different behaviour, 
    especially with respect to specific semantic reasoning patterns and 
    higher-level properties such as robustness/sensitivity with respect to target features.
    In this work, we have been able to explicitly observe previously suspected biases    
    in certain large NLI models.
    For example, previous observations~\cite{rozanova-decomposing, yanaka-med} that \textbf{roberta-large-mnli} is biased in favour of assuming upward-monotone contexts, ignoring the effects of things like negation markers, agrees with our observations that it exhibits poor context sensitivity (a low TCE influence of contexts which should be changing the output label).
    Furthermore, the causal flavour of the study adds a complimentary narrative to works that investigate model internals via probing~\cite{rozanova-decomposing} 
    and observe the presence/absence of intermediate semantic features in model representations.
    Instead of merely suggesting that these features are captured, we are able to gain insight into their causal influence via connected causal effect estimates. The causal measures presented here show us that even the highest-performing models
    systematically show a failure to adapt their predictions to changing context structure, suggesting an over-reliance on word relations across the premise and hypothesis.
    Finally, we have also added a new observation: that strategies shown to improve responsiveness to these context changes also end up increasing the \emph{robustness} to entailment-preserving word-pair insertion changes.
    
    \section{Limitations}
    Pretrained NLI models often differ in their labelling schemes (among themselves, and from dataset labelling schemes). 
    We have to the best of our ability attempted to cross-check the correctness of label configurations used, 
    but there is always the possibility in this case, and we encourage 
    reproduction and checking of the results before reporting them in any 
    external works.
    
    The causal modeling in section~\ref{sec:model_causal_diagram} draws 
    heavily on the work in~\citet{stolfo-maths}, where there is a small 
    amount of inconsistency between the textual description of their modelling 
    strategy and the labels in their diagrams. 
    We follow the version in the textual description rather than splitting the template input variable into 
    two further text variables - the ``relevant'' and ``irrelevant'' portion of the text - as such a diagram scheme would require somewhat different interventions than the ones performed. 
    We place some trust in the authors' derivations of the intervention schemes for the target causal effect measures, as we are not causal modelling experts.
    We also point out that the causal effect measures presented here are 
    sample-specific estimates, which could potentially be improved upon 
    with even larger samples of data.

\bibliography{anthology,custom}
\bibliographystyle{acl_natbib}

\appendix

\end{document}